\documentclass[conference]{IEEEtran}
\IEEEoverridecommandlockouts
\usepackage{cite}
\usepackage{amsmath,amssymb,amsfonts}
\usepackage{graphicx}
\usepackage{textcomp}
\usepackage{xcolor}
\usepackage{algorithm}%
\usepackage{algorithmicx}%
\usepackage{algpseudocode}%
\usepackage{multirow}
\usepackage{tabularx} 
\usepackage{booktabs}
\usepackage{verbatim}
\usepackage{amsmath}
\usepackage{hyperref}
\usepackage{amsfonts}
\usepackage{balance}
\usepackage{subfig}   
\def\b#1{\mathchoice{\hbox{\boldmath $\displaystyle #1$}}
        {\hbox{\boldmath $\textstyle #1$}}
        {\hbox{\boldmath $\scriptstyle #1$}}
        {\hbox{\boldmath $\scriptscriptstyle #1$}}}

\def\BibTeX{{\rm B\kern-.05em{\sc i\kern-.025em b}\kern-.08em
    T\kern-.1667em\lower.7ex\hbox{E}\kern-.125emX}}
\begin{document}

\title{High-Level Multi-Robot Trajectory Planning And Spurious Behavior Detection\\
}

\author{
\IEEEauthorblockN{
Fernando Salanova, 
Jesús Roche, 
Cristian Mahulea,  
Eduardo Montijano 
}
\IEEEauthorblockA{
Dept. of Systems Engineering and Computer Science, University of Zaragoza, Zaragoza, Spain\\
Emails: \{fsalanova, j.roche, cmahulea, emonti\}@unizar.es
}
\thanks{This work has been supported by the ONR Global grant N62909-24-1-2081, Spanish project PID2024-159284NB-I00 funded by MCIN/AEI/10.13039/501100011033, by ERDF A way of making Europe and by the European Union NextGenerationEU/PRTR, and DGA T45-23R and T64-23R.}
}

\maketitle

\begin{abstract}

The reliable execution of high-level missions in multi-robot systems with heterogeneous agents, requires robust methods for detecting spurious behaviors. 
In this paper, we address the challenge of identifying spurious executions of plans specified as a Linear Temporal Logic (LTL) formula, as incorrect task sequences, violations of spatial constraints, timing inconsistencies, or deviations from intended mission semantics. To tackle this, we introduce a structured data generation framework based on the Nets-within-Nets (NWN) paradigm, which coordinates robot actions with LTL-derived global mission specifications. We further propose a Transformer-based anomaly detection pipeline that classifies robot trajectories as normal or anomalous. Experimental evaluations show that our method achieves high accuracy (91.3\%) in identifying execution inefficiencies, and demonstrates robust detection capabilities for core mission violations (88.3\%) and constraint-based adaptive anomalies (66.8\%). 
An ablation experiment of the embedding and architecture was carried out, obtaining successful results where our novel proposition performs better than simpler representations.
\end{abstract}


\section{Introduction}

Multi-robot systems significantly enhance capabilities for complex missions such as surveillance, rescue, and area coverage by offering improved efficiency, robustness, and fault tolerance over single-agent systems \cite{DBLP}. Their ability to execute tasks in parallel and recover from individual failures makes them ideal for dynamic and high-stakes environments.

However, coordinating multiple agents introduces new challenges, particularly in ensuring that the collective behavior of the team adheres to the intended mission. Spurious behaviors may emerge due to robot faults, delays, or incorrect task sequencing, which may not be immediately evident at the level of individual agents. Such deviations can compromise mission goals, introduce safety risks, or violate temporal constraints. Detecting these spurious trajectories is essential to guarantee mission integrity in multi-robot systems.

Planning frameworks for robot teams often combine low-level motion planners (e.g., A*, RRT*, potential fields) \cite{RRT,gasparetto2015path} with high-level task models based on formal methods like Linear Temporal Logic (LTL), Petri nets (PNs), or hierarchical automata \cite{kloetzer2014petri,gujarathi2022mt}. While low-level planners focus on collision-free execution in continuous space, high-level mission planners govern the temporal and logical structure of team behavior. 
The alignment between these layers is nontrivial, making it difficult to detect team spurious behaviors, given that they can arise even if each robot's motion appears correct locally.
To address this problem, the main contributions of this work are (i) the creation of a dataset including diverse multi-robot high-level mission behaviors and (ii) a learning-based pipeline that detects spurious multi-robot trajectories, those that violate either low-level constraints (e.g., forbidden zones, speed violations) or high-level mission semantics (e.g., incorrect task ordering, temporal failures). Unlike most prior work, which focuses primarily on local agent anomalies \cite{haussermann2015novel,park2019multimodal}, our approach analyzes entire team trajectories, enabling detection of subtle behavioral inconsistencies at multiple levels.
Experiments in simulated environments show how our approach successfully identifies with high accuracy a wide range of spurious behaviors.

\section{Related Work}\label{Related Work}

Prior work on planning under logical constraints includes Singh and Saha~\cite{singh2024online}, who propose online LTL planning with reactive re-planning. Shi et al.~\cite{shi2022simulated} use simulated annealing for Boolean constraints but lack temporal expressiveness.  In contrast, our framework tightly integrates formal mission specifications, robot-level modeling, and cross-agent coordination.

Anomaly detection in robotics often focuses on low-level faults. Nandakumar et al.~\cite{chirayil2024anomaly} review anomaly types in Autonomous Robotic Missions (ARMs), but do not propose detection methods or high-level data representations. Häussermann et al.~\cite{haussermann2015novel} use spatial-temporal SOMs for plan execution in single robots. Khatib et al.~\cite{khatib2024enhancing} apply ML-based methods to detect coordination anomalies in leader-follower teams, focusing on task-specific behaviors. Zhang et al.~\cite{zhang2020framework} explore statistical detection in large-scale trajectory data, though lacking structured mission modeling.

Our approach addresses these limitations by detecting anomalies at the mission level using learned embeddings of structured symbolic trajectories. This enables end-to-end classification of multi-robot behaviors as mission-compliant or spurious, a capability underexplored in prior work.

\section{Methodology}\label{Methodology}
 
Our methodology is structured into five main phases: Initially the robot team and mission is modeled using the NWN paradigm, explained in \ref{ssec:NWN}; subsequently using the model, we generate the trajectories as explained in \ref{ssec:Dataset}, then in \ref{ssec:Low-level} low-level timing estimation is obtained through VMAS simulations. Following this, an embedding of the generated data is computed to create a more representative structure of  intrinsical behavioral patterns, as explained in \ref{ssec:encoding}, and as the final step, in \ref{ssec:learning} a Transformer-based neural network is trained on these embedded trajectories to classify them.

\subsection{Formal specification of multi-robot missions via NWNs} \label{ssec:NWN}

Achieving complex cooperative behaviors in heterogeneous multi-robot systems requires sophisticated planning mechanisms based on discrete event systems abstractions. In this work, we adopt the \emph{Nets-within-Nets} (NWN) paradigm, a hierarchical modeling formalism particularly well-suited for capturing high-level specifications and coordinating heterogeneous teams~\cite{hustiu2025multi}. 

\begin{figure}[!h]
\centering
\includegraphics[width=0.42\textwidth]{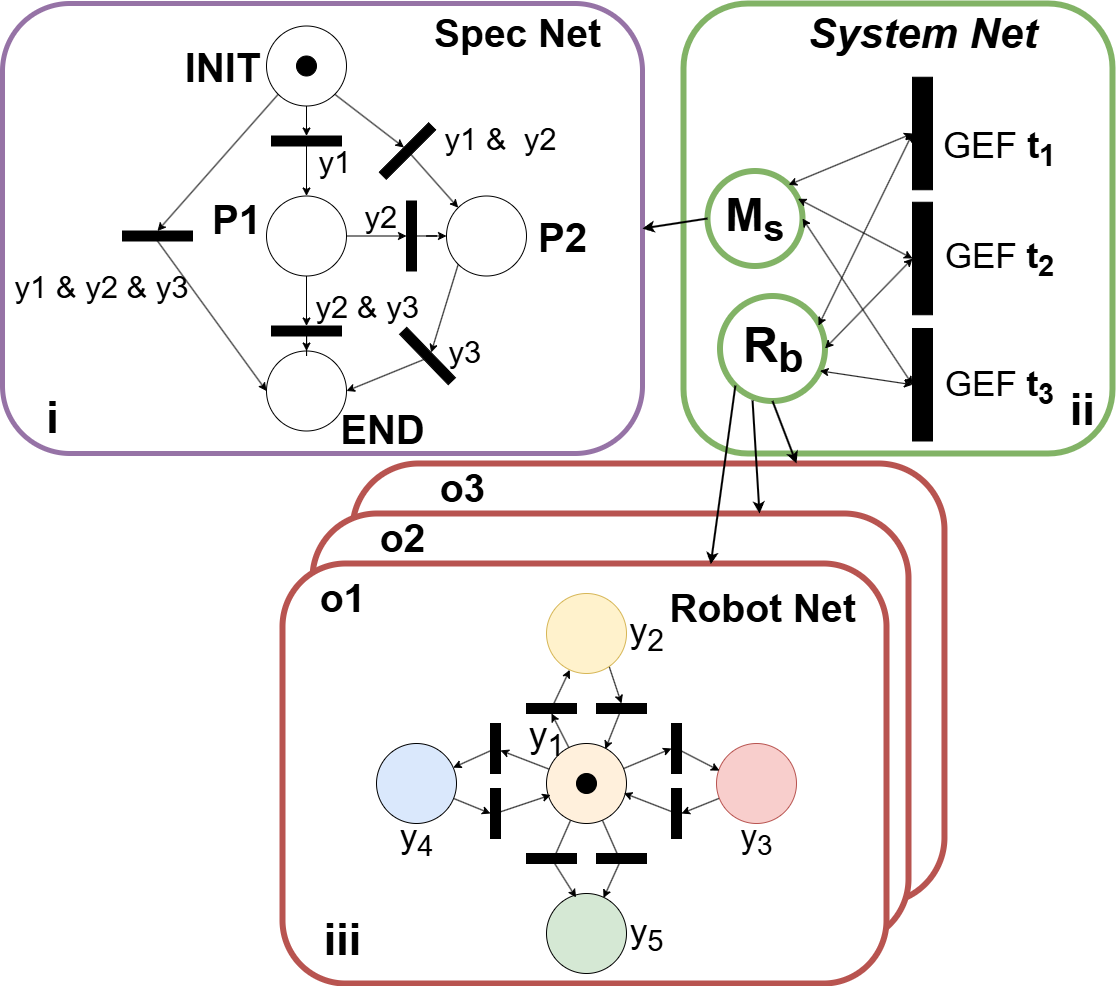}
\caption{Conceptual structure of the Nets-within-Nets framework. (i) Represents the Specification PN, in charge of ensuring a correct development of the mission. (ii) the System Net, that synchronizes the state evolution in respect to the mission and the robot actions. (iii) pictures the Robot PNs which models the agents environment and movement.}
\label{fig:nwn_structure}
\end{figure}

At its core, a NWN model of a multi-robot system consists of three main components, as illustrated in Fig.~\ref{fig:nwn_structure}. On the left, we observe the \emph{Specification Net}, a PN system  representing the global mission specification that must be fulfilled by the robot team. This mission is specified as a \emph{co-safe} LTL formula over atomic propositions, $(y_1, y_2, y_3, \ldots)$. Unlike general LTL formulas, which may require observing infinite traces, co-safe LTL formulas can be satisfied by a finite prefix of a trace.
The specification net in Fig. \ref{fig:nwn_structure} is defined by the LTL formula:
$
\lozenge ( y_1 \land \lozenge ( y_2 \land \lozenge  y_3 ) ),
$
which imposes the robot team visiting $y_1$, $y_2$, and $y_3$ in that specific order.

Below on the figure, three PNs model the behaviors of three individual robots. 
Each robot has its own \emph{Robot Net}, derived by abstracting the continuous environment into a discrete event system, for example, by a cell decomposition algorithm. 
When a robot occupies one region, the associated proposition is evaluated as true ($y_1$ in the figure), potentially enabling the firing of a transition in the specification net.

Finally, top right of the figure shows the \emph{System Net}. This high-level Petri net contains tokens that represent the previously described component nets: three tokens in place $Rb$, each of these tokens correspond to a robot net and another token in place $Ms$ corresponding to the mission specification net. 
Also for each of these transitions, a different number of robots can move synchronously, e.g., one, two or three robots moving at the same time in the example net. 
Transitions in this system net are responsible for synchronizing the evolution of the robot models with the mission specification.
The synchronization is managed via the Global Enabling Function (GEF), Java functions integrated into the system net (see \cite{hustiu2025multi} for further details).

The utility of the NWN framework for our approach lies in its ability to generate a wide range of multi-robot trajectories with inherent ground truth. By formally defining both the global mission and the individual robot behaviors, we can systematically introduce spuriousness, either as deviations from the LTL mission specification or as low-level execution anomalies during the trajectory generation process. This enables the creation of a diverse dataset containing both normal and multiple types of spurious trajectories, which is essential for training and evaluating our machine learning-based anomaly detection tool.

\subsection{Dataset generation: simulating trajectories with \emph{Renew}} \label{ssec:Dataset}

We leverage the use of the \emph{Renew} simulator \cite{kummer2004extensible} to model the NWN structure and produce detailed system traces that capture the agents’ trajectories as they follow their assigned missions. Since the specification is represented as a PN, and the NWN framework enforces its synchronized evolution alongside the robot models, the system offers formal guarantees of structural integrity and safety, thereby ensuring consistency and correctness in trajectory generation. In particular, all trajectories generated by Renew correspond to successful mission executions that satisfy the given specification.

 The algorithm for trajectory generation is formally initialized with all the above-defined Petri nets where initial markings are configured based on the selected environment. An empty buffer is then instantiated to record the generated trajectory steps. The system then enters an iterative loop that continues until a designated final place, $p_{end}$, in the Specifiaction Net ($\mathcal{N}^S$) is marked. Within each loop iteration, the system evaluates all transitions currently enabled in $\mathcal{N}^S$. For each of these, the Global Enabling Function ($GEF$) identifies all possible combinations of robot transitions that are enabled in the respective Robot nets ($\mathcal{N}^{o^j}$). A valid macro-transition is then selected, utilizing a random policy to ensure the generation of rich and varied trajectories. Firing this transition atomically updates the markings in both the $\mathcal{N}^S$ and the respective $\mathcal{N}^{o^j}$ nets. These transitions are subsequently recorded in the trajectory buffer for the current step. The execution terminates either when the $GEF$ cannot identify any possible robot transitions for the current state, signifying that the mission is impossible to complete within the given environment (e.g., the goal is completely blocked by obstacles) or when, as explained, $p_{end}$ has been reached.

Each simulation run in \textit{Renew} writes a trajectory to a plain-text
(\texttt{.txt}) file.  Every line represents a discrete timestep and lists
all robot actions that occur in parallel. Each action is recorded as a tuple,
\[
a_i = \{r_i, t_i, p_{\text{start}}, p_{\text{end}}, y_{\text{start}}, y_{\text{end}}\},
\]
where $r_i$ is the robot identifier, $t_i$ is the duration, $p_{\text{start}}$ and $p_{\text{end}}$ are the start and end places, and $y_{\text{start}}$ and $y_{\text{end}}$ denote the corresponding labels at the start and end places.
So each trajectory takes shape 
\[
T_j = \{a_1^j, ...,a_n^j\}.
\]
These symbolic logs form the raw input for the preprocessing stage, where the trajectories are converted into numerical tensors for the downstream machine-learning pipeline.
The Global Enabling Function ($GEF$) and robot-specific subnets facilitate the controlled injection of spurious behaviors, including coordination failures, timing mismatches, or task divergence. Following a similar generation process used for correct trajectories, spurious datasets are created by modifying either the robot nets to alter scenarios or introduce timing mismatches, or by subtly adjusting the specification net to induce subtle high-level behavioral deviations from the intended mission.

\subsection{Low-Level Simulation} \label{ssec:Low-level}

\begin{figure}[!h]
    \centering
    \includegraphics[width=0.31\linewidth]{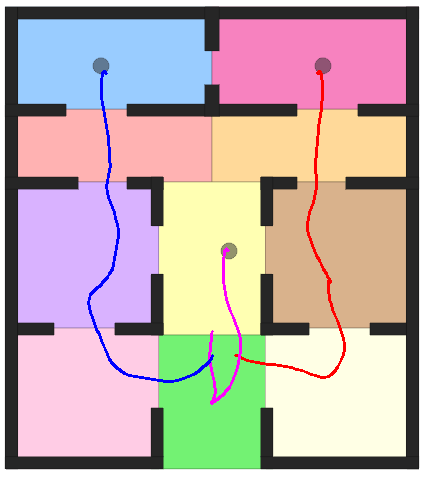}
    \caption{Visualization of designed scenario and low-level trajectories simulated in VMAS for the given example $
\lozenge ( y_1 \land \lozenge ( y_2 \land \lozenge  y_3 ) )$}
    \label{fig:VMAS_Traj}
\end{figure}

Prior to the learning stage, we introduce a simulation step to obtain realistic durations of the robot transitions between regions of interest, $t_i$, something that high-level symbolic simulations (as NWN) cannot capture accurately. For this, we leverage VMAS\cite{bettini2022vmas}, a low-level multi-robot simulator that handles continuous double-integrator dynamics, inter-robot interactions, and collision physics. Recreating task-relevant scenarios in VMAS, we extract empirical transition times between high-level places, based on physically plausible trajectories and realistic motion constraints, simulating a team of holonomic robots fulfilling a specified mission.
To bridge the semantic gap between high-level actions (e.g., moving from one place to another) and the continuous control expected by the simulator, we use a low-level policy combining RRT* for path planning and potential fields for reactive collision avoidance. Additionally, the simulator is used for visualizing robot interactions and validating temporal plans as shown in Fig.\ref{fig:VMAS_Traj}. 

\subsection{Trajectory Encoding for Machine Learning} \label{ssec:encoding}

To convert the symbolic multi-robot trajectories produced by \emph{Renew} into the fixed-size, numerical inputs required by our anomaly detection algorithm, we employ an \emph{action-token} encoding: each individual robot action becomes a dense vector, avoiding the variable-length issues of timestep- or trajectory-level encodings. Concretely, suppose we have the raw simulator record $T_j$. We parse this into five fields and embed each as follows:
Robot ID $r_i$ is mapped to a trainable embedding matrix 
\begin{equation}
\mathbf{W}_{\text{ID}} \in \mathbb{R}^{r \times d_{\text{ID}}}
\end{equation} 
where $r$ is the number of robots and we set  $d_{ID}=32$. Then, $r_i$ is embedded in the vector $\b{V}_{ID}$ equal to the $i$-th row of the matrix $\mathbf{W}_{ID}$. 
Similarly, both source and destination places share a single embedding matrix 
\begin{equation}
\mathbf{W}_{\text{place}} \in \mathbb{R}^{|P^{o}| \times d_{\text{place}}}
\end{equation}
\noindent where $|P^{o}|$ is the total number of places in each robot net and $d_{place}=32$. At initialization, $\b{W}_{place}$ is filled with random values and then trained jointly with the rest of the model. Then embeddings for $p_{\text{start}}$ and $p_{\text{end}}$ are simply the corresponding rows of $\mathbf{W}_{place}$, respectively.
The continuous action duration $t_i$ is encoded using a fixed \emph{2D angle embedding} of dimension 2. This is implemented by first applying a logarithmic normalization to the duration to compress its range, which is particularly useful for handling highly variable values. The normalized value is then scaled to an angle $\theta$ in the range $[0, 2\pi]$,
\begin{equation}
\theta = 2\pi \cdot \dfrac{\log(\texttt{\( t_i \)} + 1)}{\log(t_{\text{max}} + 1)}, 
\end{equation}
where $t_{\text{max}}$ is the maximum observed duration in the dataset. The final embedding vector is $\mathbf{V}_t = [\sin\theta, \cos\theta]^T \in \mathbb{R}^2$.
For label‐transition encoding (the origin  \( y_{\text{start}} \) and the destination \( y_{\text{end}} \)), we create a single vector
\begin{equation}
\mathbf{V}_{\text{label}} \in \{-1, 0, 1\}^{N_{\text{reg}}}
\end{equation}
where \(N_{\text{reg}}\) is the total number of atomic propositions.
\begin{equation}
v_i =
\begin{cases}
-1, & \text{if } i \text{ corresponds to the origin label} \\
1,  & \text{if } i \text{ corresponds to the destination label} \\
0,  & \text{otherwise}
\end{cases}
\end{equation}
The step index $k$ (denoted by the line where the action is within the trajectory generated with renew), for each action is encoded using a fixed, non-trainable sinusoidal positional embedding vector $\mathbf{V}_{step}\in\mathbb{R}^{d_{\text{pos}}}$, where $d_{\text{pos}} = \texttt{32}$. We interpret $\mathbf{V}_{step}$ as comprising $d_{\text{pos}}/2$ \emph{channels}, with each channel $i=0,\ldots,\tfrac{d_{\text{pos}}}{2}-1$ corresponding to one sine–cosine pair at a specific frequency. Specifically:
$$
\bigl[\mathbf{V}_{step}\bigr]_{2i}   = \sin\!\Bigl(\frac{k}{10000^{\frac{2i}{d_{\text{pos}}}}}\Bigr),
\quad
\bigl[\mathbf{V}_{step}\bigr]_{2i+1} = \cos\!\Bigl(\frac{k}{10000^{\frac{2i}{d_{\text{pos}}}}}\Bigr).
$$
Here, “channel” refers to the pair of vector entries $(2i,2i+1)$, each encoding the position $k$ at a distinct wavelength. This construction yields a vector $\mathbf{V}_{step}$ that varies smoothly with $k$, enabling the Transformer to infer the relative ordering of actions without any additional learnable parameters.
These sub-vectors are concatenated into a single action token:
\begin{equation}
\mathbf{V} \;=\;
\Bigl[
\underbrace{\mathbf{V}_{\mathrm{ID}}}_{32}
\;\Vert\;
\underbrace{\mathbf{V}_{\mathrm{src}}}_{32}
\;\Vert\;
\underbrace{\mathbf{V}_{\mathrm{dst}}}_{32}
\;\Vert\;
\underbrace{\mathbf{V}_{\mathrm{label}}}_{N_{\mathrm{reg}}}
\;\Vert\;
\underbrace{\mathbf{V}_{\mathrm{dur}}}_{2}
\;\Vert\;
\underbrace{\mathbf{V}_{\mathrm{step}}}_{32}
\Bigr]
\end{equation}
For instance, if $N_{\mathrm{reg}} = 10$, each token is 140-dimensional.

\subsection{Machine Learning Architecture } \label{ssec:learning}

\begin{figure}[!h]
\centering
\includegraphics[width=0.505\textwidth]{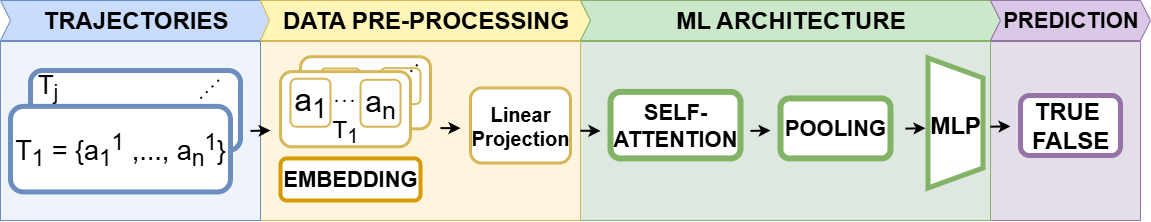}
\caption{Overview of the Transformer-based network architecture for spurious trajectory detection.(Embedding block explained in section.\ref{ssec:encoding})}
\label{fig:network_architecture}
\end{figure}

To classify trajectories as normal or spurious, we have designed a deep learning architecture to process encoded multi-robot action tokens and return classification probabilities. Figure~\ref{fig:network_architecture} synthesizes the main blocks of the architecture. 

Initially, the action tokens obtained from the encoding stage (Subsection \ref{ssec:encoding}) undergo a linear projection into a higher-dimensional latent representation with an embedding dimension set to 256 (\textbf{Embedding block} in yellow in the figure). This initial projection ensures proper scaling and structuring of the input data.
At the core of the architecture there is a Transformer encoder, composed by multiple stacked self-attention layers (green block), with the objective of identifying and prioritizing different relationships within the trajectory data.
Each layer comprises four attention heads and a Feed-Forward Network, defined by two linear transformations separated by a GELU activation function,  to further refine the feature representations generated by the attention mechanism. 
Following the Transformer encoder, the architecture employs a max pooling mechanism to aggregate the sequential output into a fixed-size representation suitable for classification. This vector is finally processed by a classifier head, implemented as a Multi-Layer Perceptron (MLP) with two fully connected layers separated by a GELU activation. The classifier employs a sigmoid activation to produce a scalar value representing the likelihood of trajectory spuriousness. 

\section{Implementation}\label{Cases of Study}

To evaluate the efficacy of our spurious high-level multi-robot trajectory detection framework, we designed and implemented a series of diverse operational scenarios. These environments were crafted to allow the generation of complex multi-robot trajectories containing a wide range of correct and anomalous high-level mission behaviors. The differences in their topological structures and functional requirements are used to highlight the capabilities of our trajectory generation and classification approach. Details found at \cite{ProjectSites}

\subsection{High-Level Behaviors and Anomalies}

The different mission cases are all derived from  co-safe LTL specifications, shown in Table~\ref{tab:LTLFormulas}. For some cases, the trajectory obtained from the spurious formulas may fall inside the normal behavior distribution (e.g., forbidden zone breaching or simultaneity violations), if this happens, the spurious trajectories are double-checked and re-labeled as true.

These are the proposed cases:
\textbf{Forbidden Zone Breaching.} The mission requires patrolling a set of places while ensuring that other designated zones are never entered. In the spurious behavior, the robot successfully patrols the required places but fails to avoid the forbidden zones.
\textbf{Sequential mission violation.} The mission defines a strict order in which the robot team must visit certain locations to complete surveillance. The spurious behavior occurs when the same locations are visited but in the wrong order.
\textbf{Mutual Exclusive Path Violation.}  The robot team is expected to reach a destination by choosing one of two mutually exclusive paths, with the condition that taking one path should completely exclude the other. In the spurious version, the team follows one path but does not fully exclude the possibility of traversing the other.
\textbf{Conditional Access Breach.} Some areas of the environment are conditionally accessible and should only be entered once specific prerequisites are met. The spurious behavior manifests when the robots access those conditional zones before satisfying the required conditions.
\textbf{Simultaneity Violation.} The mission demands to visit several locations simultaneously, requiring temporal synchronization. In contrast, the spurious behavior covers all required places sequentially rather than concurrently. 
\textbf{Recurrent Loop Modification.} The mission requires the robot team to repeat a fixed sequence of actions a limited amount of times. In the spurious behavior, the repetition is disrupted, and one of the iterations follows a modified sequence.
\textbf{Low level execution:} For these cases, the spurious behavior does not come from a direct violation of the specification, but an abnormal low level execution by one or more agents.
This behavior is obtained by modifying the transition times and place capacities (how many agents can a certain place host at the same time) of the robot net.

    \begin{table*}[htbp]
        \centering
        \footnotesize
        \renewcommand{\arraystretch}{1.5}
        \begin{tabularx}{\textwidth}{l X X}
        \toprule
        \textbf{Case} & \textbf{Normal Behavior (LTL)} & \textbf{Spurious Behavior (LTL)} \\
        \midrule
        
        \textbf{Forbidden Zone Breaching} 
        & $\lozenge(y_8 \wedge \lozenge(y_7 \wedge \lozenge y_5)) \wedge \mathcal{G}(\neg y_{10} \wedge \neg y_9 \wedge \neg y_6)$ 
        & $\lozenge(y_8 \wedge \lozenge(y_7 \wedge \lozenge y_5))$ \\
        
        \textbf{Sequentiality Violation} 
        & $\lozenge(y_1 \wedge \lozenge(y_2 \wedge \lozenge(y_3 \wedge \lozenge(y_4 \wedge \lozenge y_5))))$ 
        & $\lozenge(y_1 \wedge \lozenge(y_4 \wedge \lozenge(y_3 \wedge \lozenge(y_2 \wedge \lozenge y_5))))$ \\
        
        \textbf{Mutual Exclusion Paths} 
        & $((\lozenge y_1 \wedge \lozenge y_2 \wedge \Box(\neg y_3 \wedge \neg y_4)) \vee (\lozenge y_3 \wedge \lozenge y_4 \wedge \Box(\neg y_1 \wedge \neg y_2))) \wedge \lozenge y_5$
        & $(\lozenge y_1 \vee \lozenge y_3) \wedge (\lozenge y_2 \vee \lozenge y_4) \wedge \lozenge y_5$ \\
        
        \textbf{Conditional Access Violation} 
        & $\lozenge(y_1 \wedge \lozenge(y_2 \wedge \lozenge(y_6 \wedge \lozenge(y_7 \wedge \lozenge y_8)))) \wedge \Box((y_6 \vee y_7) \rightarrow (y_1 \wedge y_2))$ 
        & $\lozenge y_1 \wedge \lozenge y_2 \wedge \lozenge y_6 \wedge \lozenge y_7 \wedge \lozenge y_8$ \\
        
        \textbf{Simultaneity Violation} 
        & $\lozenge(y_3 \wedge y_4 \wedge y_5) \wedge \lozenge y_6$ 
        & $\lozenge y_3 \wedge \lozenge y_4 \wedge \lozenge y_5 \wedge \lozenge y_6$ \\
        
        \textbf{Incorrect Recurrency} 
        & $\lozenge (y_1 \wedge \lozenge (y_2 \wedge \lozenge y_3)) \wedge \lozenge (y_1 \wedge \lozenge (y_2 \wedge \lozenge y_3)) \wedge \lozenge (y_1 \wedge \lozenge (y_2 \wedge \lozenge y_3))$ 
        & $\lozenge (y_1 \wedge \lozenge (y_2 \wedge \lozenge y_3)) \wedge \lozenge (y_1 \wedge \lozenge (y_2 \wedge \lozenge y_3)) \wedge \lozenge (y_3 \wedge \lozenge (y_2 \wedge \lozenge y_1))$ \\
        
        \bottomrule
        \end{tabularx}
        \caption{LTL formulas for different types of explicit deviations from the mission. (Spurious trajectories that might trace correct plans are re-labeled as true (e.g. Forbidden Zone Breaching, Simultaneity Violation))}
        \label{tab:LTLFormulas}
    \end{table*}

\subsection{Training and Evaluation}

Training was conducted in a supervised manner to classify trajectories as either normal or spurious, leveraging a balanced dataset (as detailed in Section~\ref{ssec:Dataset}) to prevent class imbalance bias. The sets for each experiment contain 2000 trajectories each. 
We employed binary cross-entropy (BCE) loss and optimized the model parameters using the Adam optimizer, which is well-suited for probability-based classification tasks. Due to the intrinsic difference between the cases, the system undergoes retraining for each specific case of anomaly. This approach ensures that the model's internal parameters, or weights, are uniquely optimized to identify the nuanced patterns associated with each distinct type of spurious behavior. 
Also, as for the learned embeddings, all entries of $\b{W}_{ID}$ are initialized randomly and then optimized via back-propagation along with the rest of the network. The datasets were partitioned into standard classification subsets: 80\% for training, 10\% for validation, and the remaining 10\% for final testing. For the evaluation of the model's performance, we considered standard binary classification metrics: accuracy, measuring overall correct predictions; precision, indicating the proportion of true positive predictions among all positive predictions; recall, representing the proportion of true positive predictions among all actual positives; and the F1-score.

\section{Experimental Results and Analysis}

The results validate the effectiveness of our Transformer-based pipeline in detecting spurious behaviors across two major categories. We analyze performance case by case, highlighting the model’s ability to identify both clear mission violations and more nuanced anomalies related to execution quality, efficiency, or environmental context.

\begin{table}[h!]
\centering
\caption{Performance Metrics}
\label{tab:Case1Tab}
\renewcommand{\arraystretch}{1.1}
\begin{tabular}{|c|c|c|c|c|}
\hline
\textbf{Case} & \textbf{Accuracy} & \textbf{Precision} & \textbf{Recall} & \textbf{F1 Score} \\
\hline
Forbidden Zone & 0.996  & 0.987 & 0.999 & 0.994 \\
\hline
Sequentiality  & 0.883  & 0.836 & 0.967 & 0.906 \\
\hline
Mutual Exclusion  & 0.7833 & 0.7186 & 0.9408 & 0.815 \\
\hline
Conditional Access & 0.668  & 0.645 & 0.653 & 0.647 \\
\hline
Simultaneity & 0.826  & 0.831 & 0.894 & 0.829 \\
\hline
Recurrency & 0.807  & 0.815 & 0.835 & 0.825 \\
\hline
Low-Level & 0.913  & 0.917 & 0.941 & 0.923 \\
\hline

\end{tabular}
\end{table}

First, the results related to direct mission violations are shown in Table ~\ref{tab:Case1Tab}.
Our model demonstrates a strong capacity to identify explicit violations like forbidden zone breaching, achieving near-perfect metrics. This is because these anomalies represent a clear and unambiguous breach of a mission-critical constraint, such as entering a forbidden zone. The model also performs very well on Sequentiality Violations, with a high F1 score of 0.906. This indicates that our architecture successfully learns the precise temporal ordering required by the mission and can reliably flag deviations from it. Similarly, the performance for Simultaneity Violations and Recurrent Paths Change is robust (F1 scores of 0.829 and 0.825, respectively). These cases involve more complex structural or temporal patterns, requiring the model to capture intricate relationships between places and actions over time. 
In contrast, the model's performance is more challenged by Mutual Exclusive Path Violations and Conditional Access Breaches, achieving F1 scores of 0.815 and 0.647, respectively. The lower scores reflect the inherent complexity of these anomalies. For conditional access, the model must not only recognize locations but also understand and enforce a causal relationship, that a specific condition must be met before an action can occur. The significant drop in performance for this case underscores the difficulty in distinguishing context dependent failures from valid behavior. 
For the low-level case, the results confirm the model's strong capability in detecting anomalies related to the efficiency of execution, with F1 score of 0.923. These anomalies are characterized by clear, distinct patterns that deviate significantly from the norm, such as  deviation from a predictable timing distribution or multiple agents being in the same place at a particular timestep. 

\begin{table}[ht]
\centering
\caption{Performance Metrics for Different Embedding Methods}
\label{tab:embedding_performance}
\begin{tabular}{l c c c c}
\hline
\textbf{Method} & \textbf{Accuracy} & \textbf{Precision} & \textbf{Recall} & \textbf{F1 Score} \\
\hline

Simple Embedding  & 0.618 & 0.587 & 0.622 & 0.593 \\
Raw Embedding   & 0.513 & 0.531 & 0.554 & 0.511 \\
Mean Pooling    & 0.821 & 0.816 & 0.885 & 0.842 \\
Reduced Multi-Head    & 0.803 & 0.717 & 0.756 & 0.779 \\
Increased Multi-Head    & 0.530 & 0.543 & 0.398 & 0.421 \\
\textbf{Ours}          & \textbf{0.826}  & \textbf{0.831} & \textbf{0.894} & \textbf{0.829} \\
\hline
\end{tabular}
\end{table}

To evaluate the robustness and key components of our anomaly detection pipeline, we conducted an ablation experiment, modifying specific architectural elements and measured their impact on detection performance (Table~\ref{tab:embedding_performance}) these cases are discussed for the simultaneity violation experiment, as a case where classifying spurious behaviors is more challenging.

The baseline embedding representations are the following.
\textbf{Simpler Embedding Methods:} Replacing complex embeddings, such as learned embeddings for robot IDs and places, sinusoidal positional encodings, 2D angle representations, and label transition encodings, with One-Hot encoding, significantly degrades performance. Both accuracy and F1 score drop noticeably, underscoring the necessity of expressive feature representations to capture nuanced robot behaviors.
\textbf{Raw Trajectories as Input:} Feeding raw, unprocessed trajectory data directly into the Transformer results in the poorest performance. This highlights the critical role of our encoding pipeline in translating symbolic multi-robot actions into representations the model can effectively interpret and reason over.
\textbf{Pooling Strategies:} Changing the pooling mechanism from max pooling to mean pooling causes a slight performance drop. This indicates that while temporal feature aggregation does influence classification, the specific pooling method is not a dominant factor in anomaly detection performance.
\textbf{Reduced Multi-Head Attention:} Reducing the number of attention heads or Transformer layers leads to lower recall and F1 scores. This degradation confirms that multi-head self-attention plays a vital role in capturing the complex intra- and inter-robot dynamics necessary to detect subtle anomalous patterns. 
\textbf{Increased Multi-Head Attention:} Conversely, increasing the number of attention heads excessively can also hurt performance. When the dimensionality per head becomes too small, each head fails to capture meaningful patterns, leading to a drop in overall metrics. This suggests a careful balance is needed in configuring attention granularity.

Overall, these tests confirm that the carefully engineered embeddings, pooling mechanisms, and multi-head attention architecture are crucial for robust anomaly detection in multi-robot trajectories.

\subsection{Limitations \& Future work}

Despite its performance in detecting spurious behaviors, our current model exhibits certain limitations. Primarily, the system is not generalized across all high-level spurious behavior cases. Due to the inherent complexity and distinct characteristics of different anomaly types, the model requires retraining for each specific case, resulting in unique sets of learned weights. Conditioning the model to the LTL specification could remedy this issue.
Another limitation is the model's inability to precisely localize the spurious behavior within a given trajectory. While the system can accurately classify a trajectory as spurious after completion, it does not yet pinpoint the exact temporal segment or action where the anomaly occurs.
In the future, we plan to exploit the available tools to produce supervised data to introduce fine-grained information to extend our current solution to provide refined predictions, e.g., which robot caused the failure and when.

\section{Conclusion}\label{Conclusion}

This paper presents a model for detecting anomalies in high-level missions trajectories in multi-robot teams. Bridging the trajectory and datasets generation using discrete event systems models for ensuring correctness and security on the high-level compliance and a novel embedding, specially crafted for the generated data. Using these elements as input for the proposed machine learning architecture, successful results were achieved for multiple high-level behavior cases. When some of these cases contains complex temporal and spatial dependencies that the system could determine with high accuracy, and achieving almost perfect classification for those cases exhibiting explicit mission violations.
The proposed techniques were also compared with more naive approaches. Proving that these simpler methods lack the ability to fulfill the requirements of the proposed objective and ensuring that our model was extracting useful contrasts and relations in the data, necessary for correct high-level deviation detection.


\balance

\end{document}